%% file: llava-moe.tex
\pdfoutput=1
\documentclass[10pt,twocolumn,letterpaper]{article}

\usepackage{cvpr}              

\usepackage{times}
\usepackage{epsfig}
\usepackage{graphicx}
\usepackage{amsmath}
\usepackage{xcolor}
\usepackage{tabularx}
\usepackage{enumitem}
\usepackage{nameref}

\definecolor{cvprblue}{rgb}{0.21,0.49,0.74}
\usepackage[pagebackref,breaklinks,colorlinks,citecolor=cvprblue]{hyperref}
\usepackage{multirow}
\usepackage{amssymb}
\usepackage{pifont}
\newcommand{\cmark}{\ding{51}}%
\newcommand{\xmark}{\ding{55}}%
\usepackage{array}
\newcolumntype{C}[1]{>{\centering\arraybackslash}p{#1}}
\def\paperID{*****} 
\def\confName{CVPR}
\def\confYear{2024}

\title{LLaVA-MoLE: Sparse Mixture of LoRA Experts for Mitigating Data Conflicts in Instruction Finetuning MLLMs}

\author{Shaoxiang Chen,\quad Zequn Jie,\quad Lin Ma\\
Meituan Inc.}

\begin{document}
\maketitle
\input{sec/0_abstract}
\input{sec/1_intro}
\input{sec/2_related}

\input{sec/3_method}
\input{sec/4_exp}
\input{sec/5_concl}
{
    \small
    \bibliographystyle{ieeenat_fullname}

\input{llava-moe.bbl}
}

\end{document}

%% file: sec/0_abstract.tex
\begin{abstract}
Instruction finetuning on a variety of image-text instruction data is the key to obtaining a versatile Multimodal Large Language Model (MLLM), and different configurations of the instruction data can lead to finetuned models with different capabilities. However, we have discovered that data conflicts are inevitable when mixing instruction data from distinct domains, which can result in performance drops for tasks of a specific domain. To address this issue, we propose to apply an efficient Mixture of Experts (MoE) design, which is a sparse Mixture of LoRA Experts (MoLE) for instruction finetuning MLLMs. Within the Transformer layers, we extend the popular Low-Rank Adaption (LoRA) method by creating a set of LoRA experts specifically for the MLP layer, and route each token to the top-1 expert based on a routing function, allowing adaptive choices for tokens from different domains. Since the LoRA experts are sparsely activated, the training and inference cost are kept roughly constant compared to the original LoRA method. By replacing the plain-LoRA of LLaVA-1.5 with our MoE design, our final model is named LLaVA-MoLE. Extensive experiments proved that LLaVA-MoLE effectively mitigates the data conflict issue when mixing multiple distinct instruction datasets with various configurations, and achieves consistent performance gains over the strong plain-LoRA baselines. Most importantly, on the mixed datasets, LLaVA-MoLE can even outperform the plain-LoRA baseline trained with twice the samples.

\end{abstract}

%% file: sec/1_intro.tex
\section{Introduction}
Large language models (LLMs)~\cite{brown2020language,achiam2023gpt} have demonstrated their remarkable capabilities in following human instructions to complete various tasks, and one of the key to obtain such capability is instruction finetuning (or supervised finetuning, SFT)~\cite{wei2021finetuned}. 
Similarly, efforts have been made to create instruction-finetuned multimodal large language models (MLLMs), which connect pre-trained vision encoders with LLMs, resulting in models that are capable of answering questions given visual and textual inputs. 

\begin{figure}[t]
\centering \includegraphics[width=0.98\linewidth]{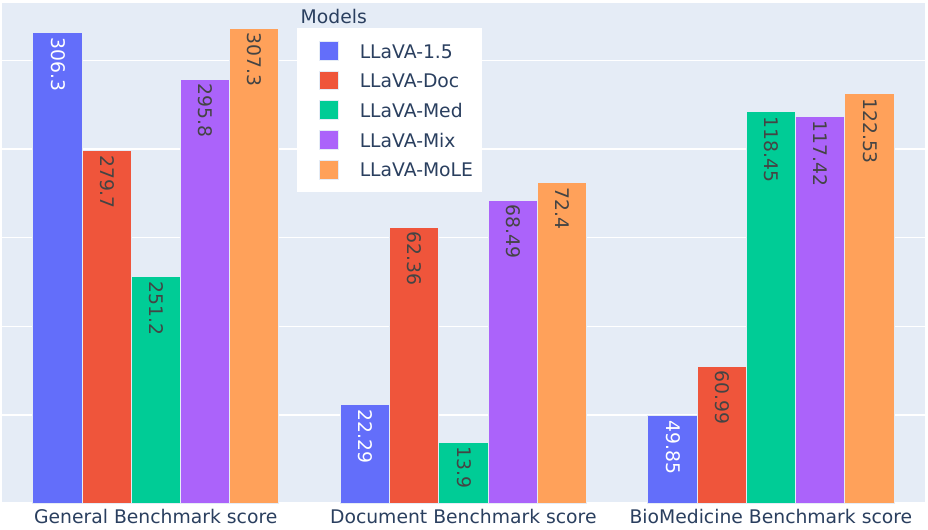}
\caption{Model performances on three benchmarks when trained with different data configurations. 
LLaVA-1.5, LLaVA-Doc, and LLaVA-Med are trained on general multi-task, document, and biomedicine datasets, respectively.
While LLaVA-Mix and LLaVA-MoLE are both trained on the mixture of all three datasets.
The performance of LLaVA-Mix the document benchmark benefits from mixing all datasets, however, the performance all other benchmarks drops after mixing. Our proposed LLaVA-MoLE successfully resolves data conflicts and maintains high performances on all benchmarks.}
\label{fig:1}
\end{figure}

Although a pre-trained LLM (7B/13B parameters)~\cite{vicuna2023,touvron2023llama} is usually included in a MLLM, the multimodal instruction training data still dominates the capability of the trained MLLMs. 
Thus a large portion of the MLLM-training effort is assigned to constructing high-quality and diverse multimodal instruction data.
For example, LLaVA-1.5~\cite{liu2023improved} carefully selected a wide range of academic task-oriented data and controlled the data size of each task. 
The resulting LLaVA-1.5 model demonstrates strong performances on benchmarks of various common vision-language tasks. 
Other successful multimodal instruction finetuning datasets~\cite{chen2023sharegpt4v,li2023m3it} are also constructed with a carefully designed data configuration. 
In addition to data, a popular and effective parameter-efficient finetuning method named LoRA (Low-Rank Adaptation)~\cite{hu2021lora} is also the key to LLaVA-1.5's success. 
LoRA reduces the number of trainable parameters of Transformers by freezing the pre-trained model weights training only an injected pair of low-rank decomposed weight matrices for each linear layer, 
which makes it faster to finetune pre-trained large models and is widely adopted in MLLM finetuning~\cite{liu2023improved,chen2023minigpt,zhang2023internlm,zhao2023svit,yin2023lamm}.

However, when data configuration is critical to MLLMs, we find in our preliminary studies that current MLLMs trained with plain LoRA are sensitive to the training data configuration.
As shown in Figure~\ref{fig:1}, we adopt three instruction tuning datasets from different domains: 1) a general multi-task dataset that contains a mixture of various vision-language instructions data, 2) a document-oriented dataset built for chart, table, and document understanding, and 3) a biomedicine dataset consists of question-answer pairs on pathology images. 
Three models are finetuned on each dataset, the resulting models are named LLaVA-1.5, LLaVA-Doc, and LLaVA-Med, respectively.
To test the finetuned model's capability on each domain, three individual benchmarks are employed\footnote{The details of these datasets and benchmarks are in Sec~\ref{sec:data_config}}.
When the MLLM is finetuned on each individual dataset, it achieves reasonable performance on the corresponding benchmark. 
But when mixing the document and biomedicine dataset with the general dataset, the trained LLaVA-Mix's performance on the general benchmark drops considerably from 306.3 to 295.8, which means there is a conflict incurred by adding data that are distinctly different from general multi-task instructions. This greatly hinders extending a MLLM's abilities by adding training data from novel domains.

To address the above mentioned issue, we propose to apply a sparse mixture of LoRA experts to LLaVA-1.5 for instruction finetuning, resulting in our proposed model named LLaVA-MoLE. 
We extend the common LoRA finetuning paradigm used by LLaVA-1.5 and many other MLLMs.
Concretely, we redesign how LoRA is applied to the MLPs in the Transformer layers of the LLM. 
Instead of adding only one pair of low-rank decomposed matrices to the original linear layer, we introduce a set of experts with the same structure as the original LoRA but different weights. 
Then for each token, these experts are sparsely routed by a router function conditioned on the token embedding, i.e., only one LoRA expert is activated and its output of the token is added to the original MLP's. 
Since the image and text tokens from different domains can exhibit distinct features, they are routed to different experts and the MLLM's ability to handle multiple domains is expanded.
In our extensive experiments on various data configurations, we discover that LLaVA-MoLE can effectively mitigate the conflicts between different instruction datasets, while maintaining roughly the same computational cost as the plain-LoRA model.
We will further show in Sec.\ref{sec:res_main} that under data conflicts, even if the plain-LoRA model is trained on twice the samples (by repeating each dataset in the mixture), its scores on the general benchmark can continue to increase but still fall behind LLaVA-MoLE. In this case, LLaVA-MoLE can achieve better performance with half the training iterations, which is a significant cost reduction.

The contributions of this paper are summarized as follows:
\begin{enumerate}
    \item Based on an advanced MLLM model and large scale datasets, we identify the data conflict issue when instruction finetuning a MLLM on a mixture of distinctly different instruction datasets.
    \item We propose LLaVA-MoLE, which is instruction-finetuned with a sparse mixture of LoRA experts to resolve the data conflict issue without significantly increasing training computation or memory. Our method further allows us to adjust the sampling ratio of each dataset in the mixture to achieve higher performance on a specific task without affecting others.
    \item Extensive experiments prove that LLaVA-MoLE achieves consistent performance gains for various data configurations on multiple benchmarks compared to using plain LoRA finetuning. 
\end{enumerate}

%% file: sec/2_related.tex
\section{Related Work}
\textbf{Multimodal Large Language Models (MLLMs).}
Current MLLMs (e.g., MiniGPT-4~\cite{zhu2023minigpt}, LLaVA~\cite{liu2023visual}, LLaVA-1.5~\cite{liu2023improved}, MiniGPT-v2~\cite{chen2023minigpt}, InstructBLIP~\cite{dai2305instructblip}, Qwen-VL~\cite{bai2023qwen}) are constructed by connecting a pre-trained vision encoder with a LLM (e.g., Vicuna~\cite{vicuna2023}, LLAMA2~\cite{touvron2023llama}). The vision encoders are usually from CLIP~\cite{radford2021learning} so that they can inherently extract semantically aligned visual features. The visual features are then adapted by a specialized light-weight module to map them into the hidden space of LLMs, so they can be jointly processed with the textual inputs by  the LLM. Through multimodal training, the MLLMs learn to generate responses given the visual and textual inputs. 
Some MLLMs are designed for domain-specific tasks by finetuning on such instruction data.
Ferret~\cite{you2023ferret}, Shikra~\cite{chen2023shikra}, and GPT4ROI~\cite{zhang2023gpt4roi} constructed referring image grounding data to train MLLMs with grounding capabilities.
mPLUG-DocOwl~\cite{ye2023mplug}, UReader~\cite{ye2023ureader}, and Vary~\cite{wei2023vary} trained document-oriented MLLMs using mixed datasets of chart, table, document, and OCR data.
However, popular MLLMs with general multi-task capabilities in domains like document understanding due to the absence of such instruction data, and as we have shown above, this can not be achieved by simply adding data.

\begin{figure*}[t]
	\centering \includegraphics[width=0.98\linewidth]{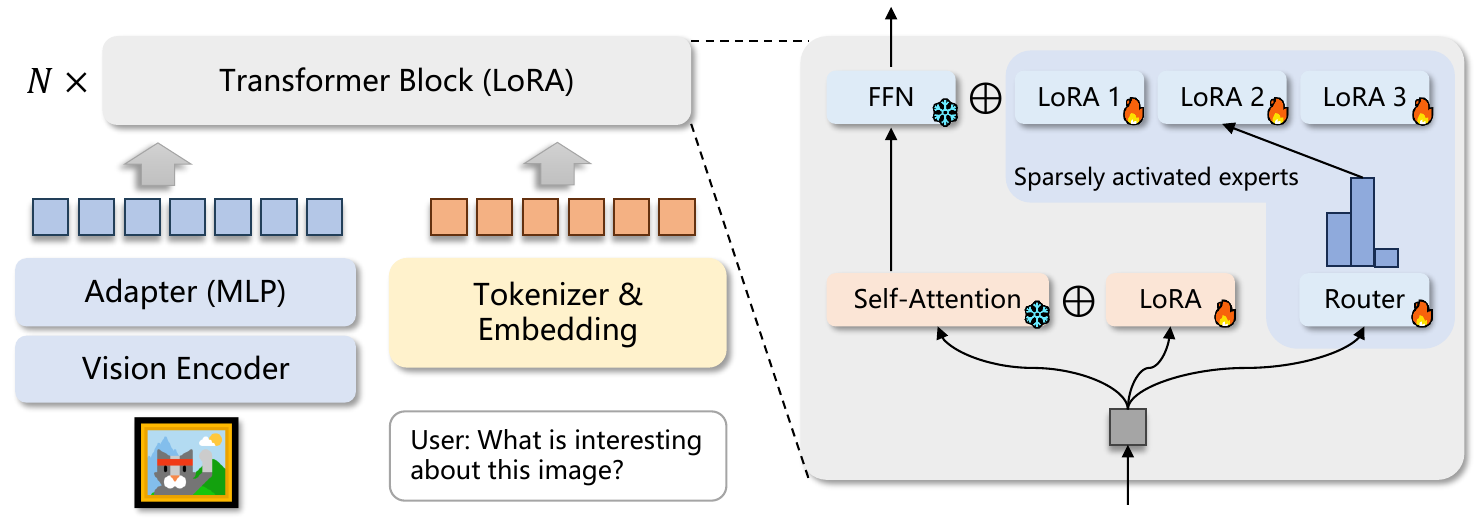}
	\caption{Overall framework of our LLaVA-MoLE with Sparse Mixture of LoRA Experts. Our model is based on LLaVA-1.5, where the input image is processed by a CLIP ViT and then projected with a two-layer MLP. The input text is tokenized and embedded, and then concatenated with the visual input to feed into the LLM. Each layer of the LLM is trained with our proposed Sparse Mixture of LoRA Experts. The FFN selects and combines with one LoRA expert according to the router's output distribution. The self-attention is also trained with LoRA but no MoE is applied.}
	\label{fig:2}
\end{figure*}

\textbf{Mixture of Experts (MoE)}, which is dynamically combining the decision of multiple experts on the same input to improve overall performance, has been studied for a long time~\cite{jacobs1991adaptive,jordan1994hierarchical}. It is gaining increasing popularity in the field of NLP~\cite{shazeer2017outrageously,lepikhin2020gshard,fedus2022switch}, where trillion-scale models can be trained with fewer computational resources, and smaller models can also be scaled to match the performance of giant models~\cite{jiang2024mixtral}. 
Recent state-of-the-art NLP models are Transformer-based, and MoE can be conveniently applied to the MLP layer of each Transformer block. 
Similarly, the idea of MoE can also be applied to scale up Vision Transformers~\cite{riquelme2021scaling}.

\textbf{Mixture of LoRA.} Since LoRA has become a successful parameter-efficient finetuning method, there has been a surge of studies to combine MoE and LoRA for more efficient and effective model tuning.
LoRAHub~\cite{huang2023lorahub} first trains several LoRA weights on upstream tasks, then to adapt to a downstream task, a gradient-free method is adopted to search for the coefficients to combine the set of pre-trained LoRA.
MOELoRA~\cite{liu2023moelora} uses a router conditioned on a task identifier to dynamically combine multiple LoRA outputs, while MoCLE~\cite{gou2023mixture} designs a router conditioned on the clustering information of each input sample. LoRAMoE~\cite{dou2023loramoe} splits the LoRA experts into two groups and explicitly learns different capabilities for each group. 
These mixture-of-LoRA methods all have predefined hyper-parameters that need to be carefully chosen, and the LoRA experts are densely mixed, i.e., by a weighted combination, which considerably increases the computational cost.
Zadouri et al.~\cite{zadouri2023pushing} compared the dense and sparse mixture of LoRA experts for large language models and concluded that a dense mixture leads to better performance. However, we will show that for instruction-finetuning MLLMs, a sparse mixture of LoRA experts can be the more economical option, i.e., it achieves comparable performances while keeping the training and inference cost roughly constant. 
Octavius~\cite{chen2023octavius} uses top-2 LoRA experts selected by a router condition on the entire input instance, which means a coarse-grained routing.
Among these works, MoCLE~\cite{gou2023mixture}, LoRAMoE~\cite{dou2023loramoe}, and Octavius~\cite{chen2023octavius} discuss the task-conflict issue, however, they studied only a few data configurations in their experiments. We will provide extensive experimental analysis for various data configurations to support our conclusions in this paper.

%% file: sec/3_method.tex
\section{Method}

\subsection{Preliminary}
\textbf{Low-Rank Adaptation (LoRA)}~\cite{hu2021lora} is an effective parameter-efficient finetuning method for Large Language Models. It can be applied to arbitrary linear layers. Formally, for a linear layer $h=Wx$ with input $x\in \mathbb{R}^{d_i}$ and weight matrix $W\in \mathbb{R}^{d_o \times d_i}$, LoRA learns a low-rank decomposed update:
\begin{equation} \label{eq1}
\begin{split}
h=Wx+\Delta Wx=Wx+ \frac{\alpha}{r} BAx,
\end{split}
\end{equation}
where $A\in \mathbb{R}^{r\times d_i}$ and $B\in \mathbb{R}^{d_o\times r}$ are the low rank matrices, $r\ll min(d_o, d_i)$ is the chosen rank, and $\alpha$ controls the magnitude of the changes to the original $W$.
During the learning of a LoRA module, only the matrices $A$ and $B$ are updated.

\subsection{Problem Formulation}
As shown in Figure~\ref{fig:2}, a MLLM can be formulated as
\begin{equation} 
\begin{split}
T^a = f_{\mathtt{MLLM}}\left( f_{\mathtt{Vis}}(I)||f_{\mathtt{Tok}}(T^q) \right),
\end{split}
\end{equation}
where $f_{\mathtt{Vis}}(\cdot)$ is the vision encoder along with the adapter that maps the input image into a sequence of visual embeddings, $f_{\mathtt{Tok}}(\cdot)$ tokenizes the input question $T^q$ and embeds the discrete tokens with a word embedding matrix, and $||$ is a sequence concatenation operation. Thus the input to the MLLM is actually a mixed embedding sequence $X\in\mathbb{R}^{L\times d}$.

The instruction data for training a MLLM is organized as triplets $(I, T^q, T^a)$, and different instruction dataset can have varying distributions, leading to different behaviors or specialties of the trained MLLM. 
We denote the $M$ instruction datasets as $\mathcal{D}_1$, $\mathcal{D}_2$, ... , $\mathcal{D}_M$.
As we have observed in Figure~\ref{fig:1}, simply mixing the instruction datasets as $\mathcal{D}_{mix}=\sum_{m=1}^{M} \mathcal{D}_m$ can cause conflicts between datasets and the MLLM can not achieve the optimal performance (compared to training on each individual dataset). 
Furthermore, different dataset mixing configurations can also lead to different model performances. 
We finally define a dataset mixture as $\mathcal{D}_{mix}=\sum_{m=1}^{M} \lambda_{m} \mathcal{D}_m$, where $\lambda_m$ represents the sampling frequency of $\mathcal{D}_m$ in the mixture.

\subsection{Sparse Mixture of LoRA Experts}
The goal of our proposed method is to mitigate the conflicts when mixing different types of instruction data. 
To this end, we introduce a set of LoRA experts and a router for each layer of the transformer. 
At each input token, the router learns to select the most suitable expert to activate, so that the model has extra capacity to handle different types of inputs. 
Assuming there are $K$ experts per layer, the expert with the highest routing function value is chosen
\begin{equation} 
\begin{split}
k=\arg\max_{j=1..K} G_j(x) = \arg\max_{j=1..K} W_{j}^{g} x,
\end{split}
\end{equation}
where $W_{j}^{g} \in\mathbb{R}^{d_i}$ is the router weight for the $j$-th expert. Then the chosen expert is activated to execute the actual computation, while the rest are simply ignored for the current token, i.e., the output of the FFN is
\begin{equation} 
\begin{split}
f'_{\mathtt{FFN}}(x) = f_{\mathtt{FFN}}(x) + E_{k}(x),
\end{split}
\end{equation}
where $f_{\mathtt{FFN}}(\cdot)$ is the original FFN module and $E_{k}(\cdot)$ is the chosen $k$-th LoRA expert, i.e.,
\begin{equation} 
\begin{split}
E_{k}(x) = \frac{\alpha}{r} B_{k}A_{k}x.
\end{split}
\end{equation}
To be more concrete, the FFN layer in modern LLMs is usually multi-layer. In this case, each linear layer of the FFN will have an individual MoE, but they share the same router, i.e., the expert choices for these layers are the same.

By only activating the top-1 expert, the actual computation cost is kept roughly the same as the original FFN with plain-LoRA. 
The extra computation comes only from the router, which is far less than the LoRA computation due to the small number of experts used in our work. 
For efficient implementation, at each layer, the input sequence $X=\{x_1,x_2,...,x_L \}$ is grouped by the expert choice of each token. For example, the sub-sequence of tokens routed to the 1-st expert  are $X^1=\{x_1^1,x_2^1,...,x^1_{L_1}\}$, where
\begin{equation} 
\begin{split}
\arg\max_{j=1..K} G_j(x^1_l) = 1,~1 \leq l \leq L_1.
\end{split}
\end{equation}
Then the computation of each $E_k(\cdot)$ can be executed in parallel for tokens in the same sub-sequence.

\subsection{Load-Balancing of Experts}
As introduced in the previous section, by routing a token to a single expert, the total computation of our MoE model is basically close to the plain-LoRA model. However, if the expert assignment is heavily imbalanced, there will be wasted idle time for the low-load experts. 

Similar to previous sparse MoE works~\cite{fedus2022switch}, we also introduce a load balancing loss for each MoE layer, which is formulated as 
\begin{equation} 
\begin{split}
\mathcal{L}_{lb}=\sum_{j=1}^{N} c_j\cdot p_j,
\end{split}
\end{equation}
where $c_j$ is the number of tokens assigned to the $j$-th expert, and $p_j$ is total routing probability of the $j$-th expert,
\begin{equation} 
\begin{split}
p_j= \sum_{x\in X} \frac{e^{G_j(x)}}{\sum_{j} e^{G_j(x)}}.
\end{split}
\end{equation}
The losses of each layer is averaged and multiplied by a constant factor $\alpha=1e^{-2}$ before added to the language modeling loss. Since the $c$ vector is non-differentiable, the gradient only flows through the $p$ vector and optimizes the router weights. 
As $\mathcal{L}_{lb}$ reduces, the expert assignment becomes closer to uniform.

Previous works set an expert capacity that ensures each expert can not process a number of tokens that exceeds the given capacity (the overflowed tokens are dropped), thus strictly limits the computation load of each expert. 
In our case, since the instruction data is relatively small compared to the text corpus used in previous works~\cite{lepikhin2020gshard,fedus2022switch}, we decide to raise the expert capacity to the maximum context length of the LLM so that no token is dropped and the experts receive sufficient training.

%% file: sec/4_exp.tex
\section{Experiments}

In this section, we present the experimental results of our proposed method on various data configurations.
\subsection{Model Architecture}
The basic model architecture follows the design of LLaVA-1.5~\cite{liu2023improved}, where a CLIP ViT-L~\cite{radford2021clip} is used as the vision encoder, with an input image resolution of 336x336 and a patch size of 14, and the adapter is a two-layer MLP that transforms the 576 tokens from the ViT. The LLM is Vicuna-7B-v1.5~\cite{vicuna2023}. During training of all our experiments, the ViT and Vicuna weights are frozen. The LoRA rank applied to the LLM is 32 if not specifically noted. 

\subsection{Training Stages and Datasets} \label{sec:data_config}
Our models are trained in two stages: pre-training and instruction finetuning. 
For the pre-training stage, we utilize the ShareGPT4V~\cite{chen2023sharegpt4v} pre-training dataset, which consists of 1.3 million detailed captioning data produced by a captioner trained on GPT4V-generated data.

For the instruction finetuning stage, we adopt multimodal instruction datasets from three different domains: general multi-task, document, and biomedicine.

M3IT~\cite{li2023m3it} and ShareGPT4V Instruct~\cite{chen2023sharegpt4v} are two general multi-task instruction datasets. M3IT collects 40 carefully curated open-source datasets and manually writes instructions for each dataset. It contains 2.4 million image-text instruction instances and we filtered its training set to 1.6 million samples to perform experiments in this paper. 
ShareGPT4V Instruct is based on the 665k LLaVA-1.5 dataset~\cite{liu2023improved}, which is gathered from publicly available task-oriented data and also conversational and complex reasoning instruction data. In addition, 23k detailed description data generated with GPT-4V~\cite{yang2023dawn} is added to form the final ShareGPT4V Instruct dataset. To evaluate general multi-task performance, we test our models on the Tiny LVLM-eHub~\cite{shao2023tiny} benchmark, which contains 42 text-related visual benchmarks, covering a wide range of tasks.

For document-oriented instruction data, we adopt the dataset collected by UReader~\cite{ye2023ureader}. It consists of image data in the form of document, table, chart, and webpage screenshot. The images and instructions are from DocVQA~\cite{mathew2021docvqa}, InfographicsVQA~\cite{mathew2022infographicvqa}, DeepForm~\cite{svetlichnaya2020deepform}, Kleister Charity (KLC)~\cite{stanislawek2021kleister}, WikiTableQuestions (WTQ)~\cite{pasupat2015wtq}, TabFact~\cite{chen2019tabfact}, ChartQA~\cite{masry2022chartqa}, TextVQA~\cite{singh2019towards}, and VisualMRC~\cite{tanaka2021visualmrc}. All these datasets are publicly available and UReader organized them into combined training and testing sets. 
We follow the data splits of UReader that contains 1.1 million resampled training instances, and we report results on the UReader's test set of ChartQA and DocVQA.
Note that the input/output length of samples in this dataset is generally longer, a considerable amount of samples can reach the maximum context length (4096) of Vicuna. 

\begin{table}[!t] 
\centering
\begin{tabular}{c|ccccc}
\toprule
Stage & Batch Size & LR & LR$_{min}$ & Warmup & MoE\\
\midrule
PT & 256 & $5e^{-2}$ & $2e^{-5}$ & 500 & \xmark\\
SFT & 64 & $2e^{-5}$ & $2e^{-6}$ & 500 & \cmark \\
\midrule
\end{tabular}
\caption{Training configurations of the pre-training (PT) and supervised instruction finetuning (SFT) stages. }
\label{table:train_param}
\end{table}

We use PathVQA~\cite{he2020pathological} as the instruction data for the biomedicine domain. It contains 32,799 questions from 4,998 pathology images. The training set has 19,755 QA pairs, and the test set has 3,370 open-ended questions and 3,391 close-ended questions. We report results on both open and close-ended sets.

For both stages, we train the models with Deepspeed ZeRO-2 optimization on 64 NVIDIA A100 80GB GPUs. Finetuning a LLaVA-MoLE model on a mixture of three datasets takes about 16 hours. The AdamW optimizer with learning rate warm up is adopted. We list important parameters of the training configuration in Table~\ref{table:train_param}.

\begin{table*}[!t] 
\centering
\begin{tabular}{c|c|C{3mm} C{3mm} C{3mm} |c|C{6mm}C{6mm}C{6mm}C{6mm}C{6mm}C{6mm}|C{10mm} C{10mm}|C{6mm} C{8mm}}
\toprule
\multirow{2}{*}{\#} &\multirow{2}{*}{Model} & \multicolumn{3}{c|}{Data Config.} & \multirow{2}{*}{MoE} & \multicolumn{6}{c|}{Tiny LVLM-eHub} & \multicolumn{2}{c|}{UReader} & \multicolumn{2}{c}{PathVQA} \\
&& $\lambda_G$ & $\lambda_D$ & $\lambda_M$ & & All & VR & VP & VKA & VC & OH & ChartQA & DocQA & Open & Closed \\
\midrule
1&LLaVA-1.5$^*$ & - & - & - & \xmark & 307.2 & 55.6 & 49.0 & 57.0 & 57.2 & 88.3 & - & - & 4.71 & 51.63 \\
2&LLaVA-Med$^*$ & - & - & - & \xmark & - & - & - & - & - & - & - & - & 38.87 & 91.65 \\
\midrule
3&LLaVA-1.5$^{\dagger}$&1 & 0 & 0 & \xmark &306.3 & 50.7 & 54.0 & 55.1 & 58.4 & 88.0 & 1.0 & 21.29 & 4.03 & 45.82 \\
4&LLaVA-Doc$^{\dagger}$&0 & 1 & 0 & \xmark &279.7 & 43.8 & 45.5 & 52.3 & 54.8 & 83.3 & 34.96 & 27.4 & 2.99 & 58.00 \\
5&LLaVA-Med$^{\dagger}$&0 & 0 & 1 & \xmark &251.2 & 37.9 & 40.3 & 38.6 & 49.2 & 85.3 & 7.56 & 6.34 & 29.28 & 89.17 \\
\midrule
6&\multirow{5}{*}{LLaVA-Mix}&1 & 1 & 0 & \xmark &298.8 & 49.8 & 49.3 & 54.8 & 58.6 & 86.3 & 36.72 & 28.26 & 3.79 & 56.85 \\
7&&1 & 0 & 1 & \xmark &299.3 & 49.8 & 49.5 & 53.9 & 59.8 & 86.3 & 10.52 & 26.11 & 30.89 & 90.17 \\
8&&1 & 1 & 1 & \xmark &297.1 & 50.0 & 50.0 & 53.0 & 59.8 & 84.3 & 37.6 & 28.34 & 30.38 & 89.09 \\
9&&1 & 2 & 0 & \xmark &290.2 & 50.0 & 49.8 & 52.1 & 54.0 & 84.3 & 40.24 & 28.54 & 3.08 & 55.97 \\
10&&1 & 2 & 1 & \xmark &292.0 & 50.8 & 50.3 & 51.4 & 55.8 & 84.3 & 39.44 & 28.66 & 29.64 & 88.58 \\
\midrule
11&LLaVA-Mix$\times$2&1 & 1 & 1 & \xmark & 295.8 & 52.8 & 47.8 & 53.3 & 59.0 & 83.0 & 40.48 & 28.01 & 28.96 & 88.46 \\
\midrule
12&\multirow{4}{*}{LLaVA-MoLE}&1 & 1 & 0 & 2 &307.3 & 52.4 & 50.5 & 57.4 & 59.6 & 87.3 & 41.36 & 30.34 & 3.05 & 58.12 \\
13&&1 & 1 & 1 & 3 &307.3 & 51.5 & 50.5 & 57.7 & 58.6 & 89.0 & 42.36 & 30.04 & 30.97 & 91.56 \\
14&&1 & 2 & 0 & 2 &310.1 & 51.9 & 52.0 & 57.1 & 60.4 & 88.7 & 44.2 & 30.3 & 3.41 & 56.23 \\
15&&1 & 2 & 1 & 3 &303.6 & 48.8 & 52.3 & 56.6 & 59.2 & 86.6 & 44.0 & 30.12 & 31.83 & 91.35 \\
\hline
\end{tabular}

\caption{Experimental results of models trained with different data and MoE configurations. $\lambda_G$, $\lambda_D$, and $\lambda_M$ are sampling frequencies for general multi-task data, document data, and biomedicine data, respectively. A sampling frequency of 0 means the dataset is not used. VR, VP, VKA, VC, and OH stands for the coarse ability categories Visual Reasoning, Visual Perception, Visual Knowledge Acquisition, Visual Commonsense, and Object Hallucination, respectively. $^*$ means the officially release model, $^{\dagger}$ indicates our reproduced models, and $\times$2 means the model is trained for two epochs. LLaVA-Mix and LLaVA-MoLE are trained with various dataset mixing configurations, we differentiate them by appending the data configuration, e.g., LLaVA-Mix[1,2,0] is experiment \#9.}
\label{table:main}
\end{table*}

\subsection{Main Results} \label{sec:res_main}
As shown in Table~\ref{table:main}, we present experimental results of models trained with different data and MoE configurations. 
We first provide results of the official LLaVA-1.5 and LLaVA-Med~\cite{li2023llavamed} models tested on each benchmark. 
For experiment \#3-5, we train plain-LoRA models individually on each dataset and name these models correspondingly as LLaVA-1.5, LLaVA-Doc, and LLaVA-Med. The performance of these models on the benchmark that corresponds to their training dataset is regarded as the baseline performance for that benchmark.
For example, Our reproduced LLaVA-1.5$^{\dagger}$ is trained specifically on the general multi-task instruction data, and it achieves an overall score of 306.3 on the Tiny LVLM-eHub, which is on-par with the official LLaVA-1.5 (307.2). And our reproduced LLaVA-Med$^{\dagger}$ achieves an accuracy of 89.17 for the closed-ended subset of PathVQA, which is close to the official LLaVA-Med's accuracy of 91.65\footnote{It is pre-trained on 600K biomedical image-text captioning data for biomedical concept alignment.}.

After the strong baselines are established, we begin to mix different datasets.
As shown by experiments \#6-8, when mixing the document instruction data and the biomedicine instruction data with the general multi-task instruction data, the overall performance of LLaVA-Mix on eHub consistently drops by 7-9 points compared to LLaVA-1.5$^{\dagger}$. 
While the UReader and PathVQA benchmark scores indicate that general multi-task instruction data is slightly beneficial for document/chart understanding and biomedical question answering, we can conclude that there are conflicts between the general multi-task data and these two types of data, and such conflicts can hurt the model's general multi-task QA abilities.

Our proposed LLaVA-MoLE can successfully resolve the above mentioned conflicts. Comparing LLaVA-MoLE[1,1,0] with LLaVA-Mix[1,1,0], we can observe that the overall performance on eHub is significantly improved to be on-par with the baseline LLaVA-1.5$^{\dagger}$, while the performance on the UReader benchmark even surpasses the baseline LLaVA-Doc$^{\dagger}$ by a significant margin, e.g., an absolute performance gain of 6.4 on ChartQA. 
This can empirically prove that the mixture of experts has learned to deal with different types of instruction data and reduce potential data conflicts.
Similarly, when we train a MoE model with 3 experts on the mixture of all 3 datasets, i.e., LLaVA-MoLE[1,1,1], the performance on each individual benchmark can surpass the corresponding baseline and the LLaVA-Mix[1,1,1].

We further adjust the data sampling frequencies for the document data and inspect the effects for LLaVA-MoLE and LLaVA-Mix. Comparing the results of experiment LLaVA-Mix[1,2,0] and LLaVA-Mix[1,1,0], when the sampling frequency of document data is increased, the performance on the UReader benchmark clearly improves as expected. However, the overall performance on eHub continues to drop (from 298.8 to 290.2). This again signifies the data conflict issue. 
Surprisingly, the performance of LLaVA-MoLE[1,2,0] on eHub is even higher than the baseline, and moreover, the improvement brought by increasing data sampling on UReader is also more significant than LLaVA-Mix[1,2,0] (e.g., a further absolute gain of 3.96 for ChartQA). 
Similar conclusions can be made when comparing results of experiments LLaVA-MoLE[1,2,1] and LLaVA-Mix[1,2,1], where the sampling frequency adjustment is performed for a mixture of 3 datasets.
These results can prove that even when the data conflict issue is amplified by adjusting the sampling frequency, our proposed LLaVA-MoLE architecture can still resolve it and achieve comparable or even higher performances on each individual benchmark. 

More importantly, if we look at LLaVA-Mix$\times$2[1,1,1], when we train the model for more epochs, in this case, each sample of these datasets is seen twice, the performance on eHub slightly improves by 3.8 but still falls behind LLaVA-Mix[1,1,1]. 
This means that the data conflict issue seriously constrains the improvement of general multi-task abilities even if more training time is consumed. 
Looking at LLaVA-MoLE[1,1,1] or LLaVA-MoLE[1,2,1], our LLaVA-MoLE models can consistently outperform LLaVA-Mix[2,2,2] by seeing less training samples. This provides a great advantage since both the training data and computational resources for MLLMs are expensive to obtain.

\begin{table*}[!t] 
\centering
\begin{tabular}{c|c|cc|c|c|cccccc|cc}
    \toprule
    \multirow{1}{*}{\#}& \multirow{2}{*}{Model}& \multicolumn{2}{c|}{Data Config.}& LoRA & \multirow{2}{*}{MoE}& \multicolumn{6}{c|}{Tiny LVLM-eHub} & \multicolumn{2}{c}{UReader} \\
    &&$\lambda_G$& $\lambda_D$ & Rank & & All & VR & VP & VKA & VC & OH & ChartQA & DocQA \\
    \midrule
    1&\multirow{8}{*}{LLaVA-Mix}&1&0&32 &\xmark&306.3 & 50.7 & 54.0 & 55.1 & 58.4 & 88.0 & 1.0 & 21.29  \\
    2&&1&1&32 &\xmark&298.8 & 49.8 & 49.3 & 54.8 & 58.6 & 86.3 & 36.72 & 28.26 \\
    3&&1&0&64 &\xmark&307.0 &53.2 &50.8 &55.3 &60.4 &87.3 &1.6 &18.8 \\
    4&&1&1&64 &\xmark&300.2 &50.8 &48.3 &55.3 &58.2 &87.6 &39.24 &29.64 \\
    5&&1&0&96 &\xmark&307.8 &51.7 &50.3 &56.4 &61.4 &88.0 &1.6 &10.46 \\
    6&&1&1&96 &\xmark&301.1 &51.1 &48.3 &54.6 &60.2 &87.0 &39.96 &29.48 \\
    7&&1&0&128 &\xmark&309.8 &53.2 &50.7 &56.4 &61.2 &88.3 &12.0 &11.3 \\
    8&&1&1&128 &\xmark&310.2 &54.6 &51.2 &56.4 &59.2 &88.6 &40.72 &30.54 \\
    \midrule
    9&\multirow{2}{*}{LLaVA-MoLE}&1&1&32 &2&307.3 & 52.4 & 50.5 & 57.4 & 59.6 & 87.3 & 41.36 & 30.34 \\
    10&&1&1&128 &2&313.6 &54.1 &49.7 &59.3 &61.8 &88.6 &45.32 &32.62 \\
    \midrule
\end{tabular}
\caption{Experimental results of models trained with different LoRA ranks. Similar to Table~\ref{table:main}, the models here can be referred to by its configuration, e.g., \#1 is LLaVA-Mix[1,0]-R32, where R32 indicates a LoRA rank of 32.}
\label{table:rank}
\end{table*}

\begin{table*}[!t] 
	\centering
	\begin{tabular}{c|cccccc|cc}
		\toprule
		\multirow{2}{*}{MoE} & \multicolumn{6}{c|}{Tiny LVLM-eHub} & \multicolumn{2}{c}{UReader} \\
		& All & VR & VP & VKA & VC & OH & ChartQA & DocQA \\
		\midrule
		2 &307.3 & 52.4 & 50.5 & 57.4 & 59.6 & 87.3 & 41.36 & 30.34 \\
		3 &303.2 & 49.4 & 50.3 & 56.9 & 58.4 & 88.3 & 41.64 & 30.2 \\
		5 &311.6 & 52.6 & 55.0 & 56.6 & 58.4 & 89.0 & 41.88 & 30.82 \\
		8 &307.3 &51.1 &52.8 &57.1 &60.0 &86.3 &40.96 &30.07 \\
		16 &306.8 & 52.6 & 51.3 & 56.0 & 59.6 & 87.3 & 42.2 & 30.48 \\
		\midrule
	\end{tabular}
	\caption{Experimental results of LLaVA-MoLE models trained with different numbers of experts.}
	\label{table:num_expert}
\end{table*}

\subsection{Ablation Studies}
\textbf{LoRA Rank.} We first inspect the effect of LoRA rank under our data and MoE configurations, and the results are shown in Table~\ref{table:rank}. As can be observed, for a LoRA rank of 32, 64, and 96, mixing the document instruction data with the general multi-task instruction data all leads to performance drop on the eHub benchmark. But comparing the results of experiments LLaVA-Mix[1,1]-R32, LLaVA-Mix[1,1]-R64, and LLaVA-Mix[1,1]-R96, we also find that increasing the LoRA rank, i.e., increasing the model capacity, can mitigate the data conflict issue to some extent: the overall score on eHub increased from 298.8 (R32) to 301.1 (R96). Moreover, if the LoRA rank is increased to 128, this issue seems to be resolved (comparing the eHub scores of LLaVA-Mix[1,0]-R128 and LLaVA-Mix[1,1]-R128). 
However, we argue that simply raising the model capacity is an expensive solution, leading to computation and memory increase during training. 
While our proposed LLaVA-MoLE can resolve this issue without incurring much extra cost. 
It is noteworthy that for both small (32) and large (128) LoRA ranks, LLaVA-MoLE outperforms LLaVA-Mix by a significant margin on both benchmarks.
Finally, comparing experiment LLaVA-MoLE[1,1]-R32 with LLaVA-Mix[1,1]-R64 (or LLaVA-Mix[1,1]-R96), where the latter has the same or even larger number of parameters, we can confirm that the effectiveness of LLaVA-MoLE is not simply brought by increasing model capacity with MoE.

\textbf{Number of Experts.} We also study the effect of the number of experts by training a series of LLaVA-MoLE models with the expert number ranging from 2 to 16, and the results are shown in Table~\ref{table:num_expert}. Note that for these experiments, we mix the general multi-task and document instruction datasets and test the models on the two corresponding benchmarks. We can see that as the expert number increases, the overall performance also improves. Using 5 experts achieves the best performance for this data configuration. If the expert number continues to increase to 8 or 16, the performance slightly drops. This could be because each expert can not receive enough training when the tokens are distributed among 8 or 16 experts. To summarize, the model performance is not very sensitive to the number of experts, and setting a small number of experts can already achieve performance advantages over non-MoE models on a million-scale dataset.

\textbf{Sparse vs. Dense MoE.} Dense MoE strategy is widely adopted by previous works of LoRA MoE~\cite{huang2023lorahub,liu2023moelora,gou2023mixture,dou2023loramoe,zadouri2023pushing}, thus we also compare sparse and dense MoE in our settings.
As shown in Table~\ref{table:vs_dense}, on a dataset mixture of the general multi-task data and the biomedicine data, sparse and dense MoE achieves similar performances on all benchmarks and both of them resolves the data conflict issue, i.e., they achieve a score of 312.6 (\#2 and \#3)on eHub compared to 299.3 from the baseline (\#1).
However, the dense MoE model consumes 83\% of the GPU memory, which is significantly more than the sparse MoE model's ratio of 61\%.
When we try to run experiments with a dense mixture of 3 experts, an out-of-memory (OOM) error is encountered on the GPU (\#5) under long input/output length. Thus it is difficult to scale-up dense MoE even for a LLM of 7B parameters\footnote{Model or tensor parallelism is not used for all of our experiments, but they do not affect the overall memory consumption}.
We would recommend using our proposed MoE architecture (scales easily to 16 experts as shown in Table ~\ref{table:num_expert}) for better scalability.

\begin{table*}[!t] 
	\centering
	\begin{tabular}{c|ccc|c|cccccc|cc|cc}
		\toprule
		\multirow{2}{*}{\#}&\multicolumn{3}{c|}{Data Config.} & \multirow{2}{*}{MoE}& \multicolumn{6}{c|}{Tiny LVLM-eHub} & \multicolumn{2}{c|}{UReader} &\multicolumn{2}{c}{PathVQA} \\
		& $\lambda_G$ & $\lambda_D$ & $\lambda_M$ && All & VR & VP & VKA & VC & OH & ChartQA & DocQA & Open & Closed \\
		\midrule
		1&1&0&1&\xmark&299.3 & 49.8 & 49.5 & 53.9 & 59.8 & 86.3 & 10.52 & 26.11 & 30.89 & 90.17 \\
		2&1&0&1&Dense, 2&312.6 &50.5 &52.7 &57.1 &63.2 &89.0 &19.52 &28.24 &32.13 &92.03 \\
		3&1&0&1&Sparse, 2&312.6 &51.6 &51.8 &58.1 &62.4 &88.6 &19.36 &28.63 &32.43 &92.01 \\
		\midrule
		4&1&1&1&\xmark& 297.1 & 50.0 & 50.0 & 53.0 & 59.8 & 84.3 & 37.6 & 28.34 & 30.38 & 89.09\\
		5&1&1&1&Dense, 3&OOM &- &- &- &- &- &OOM &OOM &OOM &OOM\\
		6&1&1&1&Sparse, 3& 307.3 & 51.5 & 50.5 & 57.7 & 58.6 & 89.0 & 42.36 & 30.04 & 30.97 & 91.56 \\
		\midrule
	\end{tabular}
	\caption{Experimental results of models trained with dense MoE and sparse MoE (LLaVA-MoLE) on different data configurations.}
	\label{table:vs_dense}
\end{table*}

\begin{figure}[t]
	\centering \includegraphics[width=0.98\linewidth]{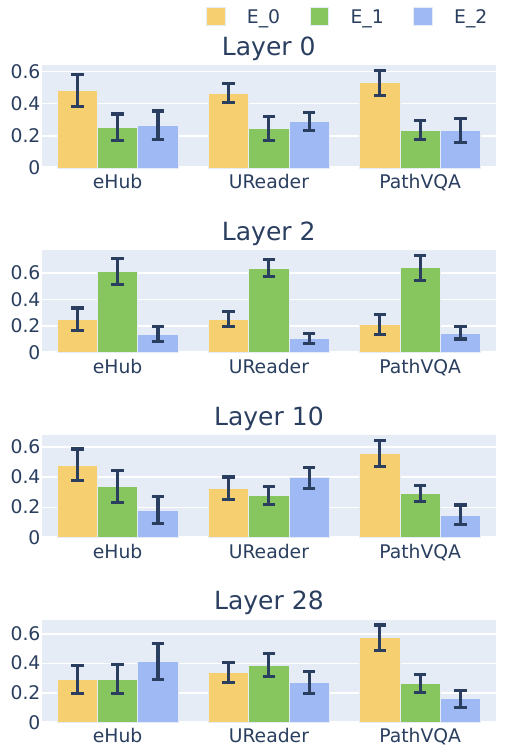}
	\caption{Average proportion of tokens assigned to each expert on different benchmarks for LLM layers 0, 2, 10, and 28. Standard deviation is shown as the error bar. E\_i represents the i-th expert. }
	\label{fig:3}
\end{figure}

\subsection{Routing Choice Visualization}
We perform a rough analysis on the routing choice of our LLaVA-MoLE model with 3 experts trained on the mixture of all three datasets.
We count the expert choices on the token sequences from each benchmark, and compute the mean and standard deviation of the proportion of tokens assigned to each expert. The results of layer 0, 2, 10, and 28 are visualized in Figure~\ref{fig:3}.
For some layers, e.g., layer 2 and 10, the expert choice patterns are similar for different types of data, but differ among layers.
There are also layers (10 and 28) where each type of data has its own expert choice pattern.
We do not observe an obvious pattern that shows a specific expert is consistently favored over the others. 
But some experts can have a slight tendency to be selected more often than the others on a specific dataset, e.g., expert 0 is activated more often on PathVQA samples across all layers. 

%% file: sec/5_concl.tex
\section{Conclusion}
In this paper, we first identified the data conflict issue when instruction finetuning multimodal large language models on a mixture of datasets from multiple distinct domains.
To address this issue, we propose LLaVA-MoLE, which uses a sparse mixture of LoRA experts to improve the plain-LoRA architecture. It uses a set of LoRA experts for the MLP layers and routes each token to the top-1 expert. Since only the selected expert is activated to execute computation, the actual computational cost for the entrie model is kept roughly the same as a normal LoRA model. In the meantime, our LLaVA-MoLE effectively mitigates the data conflict and achieves a consistent performance improvement over the plain-LoRA baselines on a variety of data configurations. 
We further verified that LLaVA-MoLE performs similarly with a dense MoE model while requiring significantly less computational resources, which is particularly advantageous for samples with long context length.
For our future work, it would be interesting to apply our method to the multi-task pre-training stage of the MLLMs, where a much larger number of training examples from multiple domains are mixed.